# A New Multiple Max-pooling Integration Module and Cross Multiscale Deconvolution Network Based on Image Semantic Segmentation

Hongfeng You, Shengwei Tian, Long Yu, Xiang Ma, Yan Xing and Ning Xin

*Abstract*—To better retain the deep features of an image and solve the sparsity problem of the end-to-end segmentation model, we propose a new deep convolutional network model for medical image pixel segmentation, called MC-Net. The core of this network model consists of four parts, namely, an encoder network, a multiple max-pooling integration module, a cross multiscale deconvolution decoder network and a pixel-level classification layer. In the network structure of the encoder, we use multiscale convolution instead of the traditional single-channel convolution. The multiple max-pooling integration module first integrates the output features of each submodule of the encoder network and reduces the number of parameters by convolution using a kernel size of 1. At the same time, each max-pooling layer (the pooling size of each layer is different) is spliced after each convolution to achieve the translation invariance of the feature maps of each submodule. We use the output feature maps from the multiple max-pooling integration module as the input of the decoder network; the multiscale convolution of each submodule in the decoder network is cross-fused with the feature maps generated by the corresponding multiscale convolution in the encoder network. Using the above feature map processing methods solves the sparsity problem after the max-pooling layer-generating matrix and enhances the robustness of the classification. We compare our proposed model with the well-known Fully Convolutional Networks for Semantic Segmentation (FCNs), DecovNet, PSPNet, U-net, SgeNet and other state-of-the-art segmentation networks such as HyperDenseNet, MS-Dual, Espnetv2, Denseaspp using one binary Kaggle 2018 data science bowl dataset and two multiclass dataset and obtain encouraging experimental results.

This research is partially supported by Xinjiang Uygur Autonomous Region Natural Science Fund Project (2016D01C050), and Xinjiang Autonomous Region Science and Technology Talent Training Project (QN2016YX0051). We would also like to thank our tutor for the careful guidance and all the participants for their insightful comments.(Corresponding author: Shengwei Tian).

The authors are with Xinjiang University, 830000 Urumqi.
H. F. You is with the School of Information Science and Engineering, Xinjiang University, 830000, China (e-mail: 1053109177@qq.com).
S. W. Tian is with the Software College, Xinjiang University, 830000, China (e-mail: tianshengwei@163.com).
L. Yu is with the Network Center, Xinjiang University, 830000, China (e-mail: yul_xju@163.com).
X. Ma is with Department of Cardiology, The First Affiliated Hospital of Xinjiang Medical University,, 830011, China (e-mail: maxiangxj@163.com).
Y. Xing is with the Imaging Center, the First Affiliated Hospital of Xinjiang Medical University, 830011, China (e-mail:xingyanzwb@sina.com).
N. Xin is Assistant Professor of Artificial Intelligence at Institute of Semiconduc- tors Chinese Academy of Sciences, 100083 ,(e-mail:ningxin@semi.ac.cn).

## I. INTRODUCTION

Currently, convolutional neural networks (CNN) are widely and very successfully used in a variety of scenarios, including image classification [1][2][3], target detection [4][5], image inpainting [44], image semantic segmentation [6][7][9][10], scene behavior prediction [11][12], video target tracking [13] [14], etc. Medical images have the characteristics of complex and diverse shapes, large individual differences, uneven signals, edge paste and multiple types of noise. Image pixel segmentation has become an indispensable part of medical image target recognition. Researchers solve the problem of segmentation of various medical images by extracting various features [6][15][16][18]. Specific features have been proposed for segmenting specific medical images, but the effectiveness of applying these features to other medical images is often poor. This situation makes it very difficult to implement automatic segmentation of medical images.

The traditional semantic segmentation of a medical image is mainly conducted by extracting the low-level features of the image. Such features [34] not only incorporate a large number of personal prior conditions but also require a significant amount of manpower and effort according to the descriptions of methods, such as Ostu [22], fuzzy c-means (FCM) [29], watershed [24] and normalized cut (N-Cut) [25]. With the increase in computer processing power, machine learning methods began to be used for image semantic segmentation. One type of method is supervised learning, such as decision trees [26], naive Bayesian [28] and support vector machine (SVM) [27]. The other type of method is unsupervised learning, performed by a clustering algorithm [29], and principal component analysis (PCA)[30]. With the emergence of the FCN [31] algorithm proposed by J Long et al, deep learning has officially entered the field of image semantic segmentation. FCN is the first deep learning algorithm to implement end-to-end dense prediction, i.e., to predict all pixels in an image. This model overcomes the limitations of manual features and has outperformed the traditional segmentation methods in terms of speed and accuracy. However, because max-pooling and upsampling layers reduce the resolution of the feature map, the latter becomes coarse. In the following studies, the researchers used a variety of network structures to improve the resolution of feature maps, such as VGG16 [32], DeconvNet [33], PSPNet [20], SegNet [34] and U-Net [35]. In recent years, some multi-input models such as HyperDenseNet[40], MS-Dual[41], Espnetv2[43], Denseaspp[42] have achieved better results.

The above models improved the resolution of feature maps to a certain extent, but some aspects can be further enhanced to obtain a better resolution. We use two new strategies to enhance the segmentation of medical images. Our main contributions are as follows:
1) We propose a multiple max-pooling feature integration strategy. A convolution kernel with the size of 1 can integrate feature maps and reduce the number of parameters. At the same time, different max-pooling layers (with different pool-sizes) are added after each convolution to achieve the translation invariance of feature maps of each submodule.
2) In the decoder network, we propose a cross multiscale deconvolution strategy. The strategy uses the feature maps of the encoder network to do the prior knowledge of convolution in the decoder network of different scales to realize the cross-fusion of the feature maps. Experiments show that this strategy is meaningful for semantic segmentation.
3) To enhance the readers' understanding of the paper, we visualize each model and analyze its advantages and disadvantages and the reasons for the visualization results in detail.

The remainder of this paper is organized as follows. In Section II, we introduce in detail some of the most popular models and some state-of-the-art segmentation networks to describe their advantages. The proposed model and the related details are presented in Section III. The experimental results obtained on simulated and real-world data and a discussion are presented in Section IV. Finally, our conclusions are given in Section V.

## II. RELATED WORK

End-to-end models have been actively applied in image semantic segmentation. We summarize some of the classic and the state-of-the-art end-to-end models as follows, and analyze their advantages and disadvantages.

The advantage of Fully Convolutional Networks for Semantic Segmentation (FCNs) [31] is that it does not perform preprocessing and postprocessing, it overcomes the limitation of manual features, and it enables fast and accurate reasoning. However, FCNs has two limitations: 1) small samples are often ignored, and 2) the feature map generated by the encoder network is too coarse, and the fully connected layer of the decoder network is too simple, resulting in the detailed structure of the segmented target often being lost and not smooth. The main contribution of VGG16 [32] is to study the depth of the network, which proves that a deeper network can improve the accuracy of image recognition. A deep network also improves the quality of the feature map. Another contribution is the comprehensive evaluation of the filters in the convolutional layer. However, this method still uses a fully connected layer in the decoder part of the network, resulting in a loss of some details. Deconvolution network [33] inherits the advantages of the VGG16 encoder network and improves the structure of the decoder network. The latter consists of unpooling, convolutional and ReLU layers and is named the deconvolution layer. The deconvolution layer uses multiple filters (multiple submodules) to convert the sparse feature map generated by unpooling into a dense feature map, thus improving the resolution of the feature map in the decoder. However, feature map extraction by the encoder network is limited; therefore, applying this approach to complex segmentation tasks is difficult. PSPNet [20] is to use a ResNet [19] network as its encoder network, which retains a higher-quality feature map. PSPNet subsequently proposes a pyramid pooling strategy that extracts the global semantic information through the average pooling layer of different scales. However, PSPNet does not use deconvolution in its decoder network, so the result of image segmentation is not smooth. In this method's pyramid pooling strategy, except for the last submodule in the encoder network, a priori information of other submodules and the original image remains unused. SgeNet [34] also uses VGG16 as its encoder. In the decoder of SgeNet, an upsampling layer is used instead of an unpooling layer, and the feature map is directly copied and expanded. At the same time, convolution is used to generate a dense feature map, which improves the resolution of the latter. Because SgeNet discards the fully connected layer, the greatly reduced, so the network can complete tasks in real time. The main idea of U-Net [35] is that feature maps of each layer in the decoder network and feature maps of each layer of the encoder network are spliced through the concatenation layer in the decoder to improve the resolution of the output feature maps. This fusion strategy can more effectively use the original deep feature maps in the encoder network while decoding, resulting in good training results for small data samples.

Recently, some state-of-the-arts multi-channel semantic segmentation methods achieved better results. Mehta S[] et al proposed a lightweight network ESPNet v2, which not only obtains a larger receptive field through expansion convolution, but also reduces the calculation of parameters.

At the same time, it integrates the deep features of expansion convolution mining with the original data features to obtain more deep features. Sinha A [] et al. proposed a Multi-scale guided attention model to realize medical image segmentation. Firstly, multi-scale features of the image are extracted by multiple ResNets, secondly, it uses guided attention to realize the deep feature mining and fusion by multi-scale features and each ResNet feature extraction result, and finally realizes the modeling. Dolz J[] et al proposed a HyperDenseNet model, which firstly mines the deep features of each channel through convolution, then fuses multiple features according to different order for the results of multiple deep features, and further excavates the fusion features by the way of Cropping to obtain fewer but more meaningful features. Yang M[] et al made intensive feature enhancement with the Atrous Spatial Pyramid Pooling (ASPP) strategy proposed by previous researchers, and used expansion convolution to mine more features through the Dense network, thus improving the feature learning ability of the ASPP model.

Considering the above research, we clearly observe that the purpose of the above models is to solve the problem of coarse feature maps to obtain more precise feature maps and to improve the detection efficiency. We summarize the following three points to solve the problems described above: 1) if more encoder feature maps are retained to provide more prior information for the decoder network, the resolution of the final feature maps can be improved such as multi-input feature parallel mining; 2) a deconvolution strategy can improve the resolution of the feature map by generating dense feature maps, and 3) the concatenation or addition strategies can improve the resolution of the final feature map by splicing the feature maps generated in the encoder network. At the same time, the fully connected layer is discarded or a convolution with a kernel size of 1 is used to reduce the operation parameters and to complete the segmentation task quickly.

In this paper, we propose a new deep end-to-end model, called MC-Net, to improve the resolution of feature maps. As shown in Figure 1, the structure of this model mainly consists of four parts: an encoder network, a multiple max-pooling integration module, a decoder network and a pixel-level classification layer. In the encoder network part, we use five layers of multiscale convolution. After each layer, we use a concatenate layer to implement the stitching of the feature map and use max-pooling to preserve the translation invariance of features, so more feature maps in the previous layer are retained for the next layer. The multiple max-pooling integration module is composed of convolutional networks with kernel sizes of 1 and max-pooling layers of different scales. The main function of this module is to integrate the feature maps of the encoder network and retain the translation invariance of the original feature maps. Finally, stitching of multilayer feature maps is performed by the concatenate layer to improve the resolution of feature maps. In the decoder network, we generate a sparse matrix from the output of the feature integration module through an upsampling layer, combine it with multiscale convolution to form multiscale deconvolution, and generate dense feature maps. The dense feature maps and the feature maps of the encoder network are added by the addition layer (note that we propose a new fusion strategy, as detailed in part D of the MC-Net section) by increasing the number of features in the channel to achieve accurate segmentation by the algorithm. Finally, we predict the output of the decoder network by using a pixel-level convolutional layer. In the experiments, the proposed model and several classic models are evaluated while using the same number of parameters on three sets of open datasets (the Kaggle 2018 data science bowl dataset, the CHAOS dataset, and the BraTS 2018 dataset), and the results indicate that our proposed algorithm has the best effect.

III. MC-NET

This section is divided into four parts. Part A mainly introduces the structure of the MC-Net model. In Part B, we introduce the encoder network of the MC-Net network in detail. In Part C, we introduce the integration module based on concatenating features generated by multiple max-pooling layers. In Part D, we propose a decoder network (a multiscale deconvolution network with feature maps cross) corresponding to the submodules of the encoder network.

*A. Model framework*

As shown in Figure 1, the MC-Net structure has three stages. In stage 1, we convert the image size to $256 \times 256$. The converted image is used as the model input. Stage 2 includes three parts of MC-Net: the encoder network for feature extraction, the multiple max-pooling integration module for invariance transfer and information integration of feature maps from each submodule of the encoder network, and the decoder network for dense feature map generation (feature map refinement). In stage 3, each pixel in the feature map is classified by pixel-level convolution.

*B. Encoder Network*

The encoder network is composed of many submodules, each of which is composed of three kinds of convolution kernels of different sizes. Different types of convolution kernels can generate a variety of feature maps, which provide more prior information for the subsequent operational modules. Each convolution uses the nonlinear ReLU activation function. Due to the ReLU activation function, all the negative features will not be activated, and the network becomes sparse; the computational efficiency is thus improved. A batch normalization (BN) layer [21] is used to normalize the convolutional feature maps to reduce the distribution influence of the weight initialization value and increase the training speed. Then, the three kinds of feature maps after normalization are stitched together by the concatenate layer, and more prior information is retained. Finally, we perform a pooling process with a max-pooling of size 2. Each pooling compresses the feature graph to reduce the network complexity and the amount of computation and remove redundant information. Each submodule has the same structure. All the equations for each of the submodules in the encoder network are as follows:

$$Con_k(i,j) = \sum_{f=1}^{n}(X * W_f)(i,j) + b \quad (1)$$

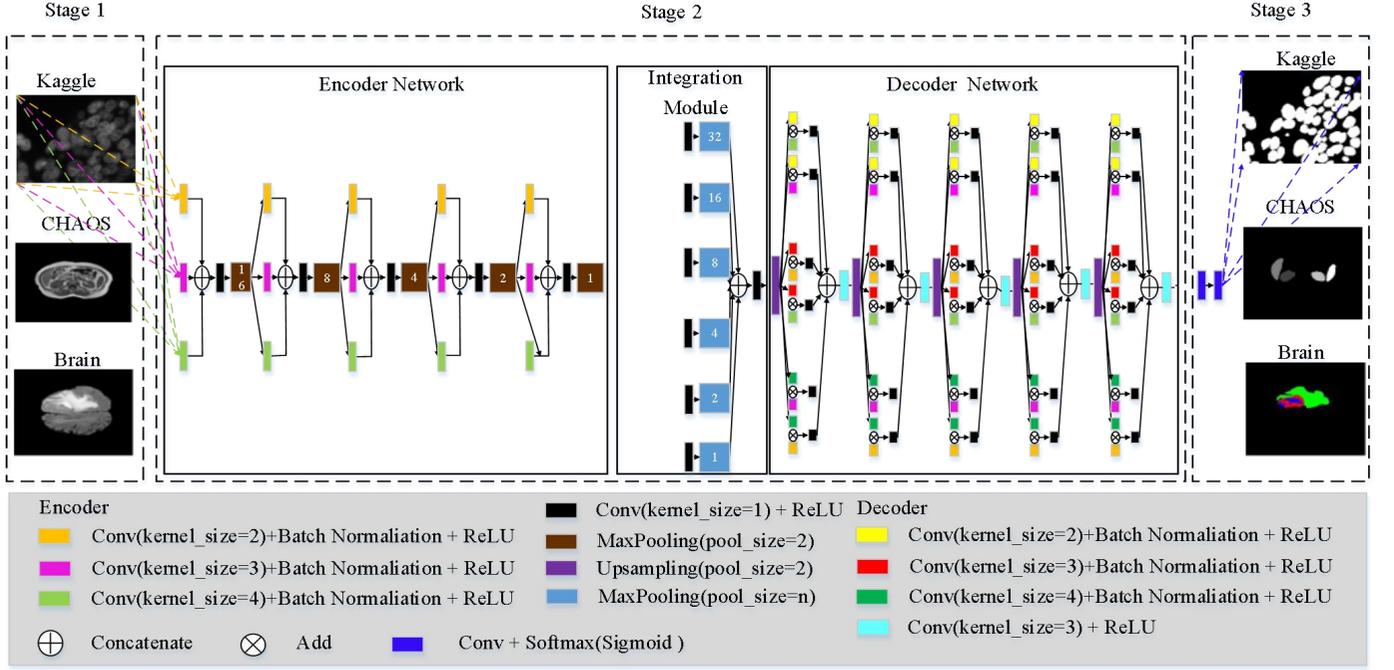

Fig. 1. Illustration of the MC-Net model. The "Encoder Network" consists of five submodules, each of which performs multiscale convolution, and all brown rectangles represent max-pooling layers where the pool-size (window size) is 2, and the numbers correspond to those in the "Integration Module". The numbers in the light blue rectangles in the "Integration Module" represent the pool-size of various max-pooling layers. "Number 32" represents the max-pooling layer that will directly process the original image, and its pool-size is equal to 32. The number of layers in the "Decoder Network" and the "Encoder Network" correspond to each other. The information of feature maps is increased by the add layer, and the multiscale feature map is spliced by the concatenate layer. The final decoder network output feature maps are fed to a sigmoid or softmax classifier for pixel-wise classification.

where $Con_k(i,j)$ represents a two-dimensional convolution, $f$ represents the number of filters, $X$ represents the input image matrix, $W$ represents the weight matrix, $k$ represents the size of the convolution kernel, $b$ represents the bias terms, $i$ represents the transverse coordinates of the image matrix, and $j$ represents the longitudinal coordinates of the image matrix.

$$ReLU_k(i,j) = Max(0, Con_k(i,j)) \quad (2)$$

where $ReLU_k(i,j)$ represents the ReLU activation function, and $Max$ represents the maximum operator.

$$BN_k(i,j) = \frac{ReLU_k(i,j) - E[ReLU_k(i,j)]}{\sqrt{Var[ReLU_k(i,j)]}} \quad (3)$$

where $BN_k(i,j)$ represents normalization operations, $E[]$ represents the average of the input image matrix, and $\sqrt{Var[]}$ represents the standard deviation of the input image matrix.

$$Output(i,j) = Concatenate(BN_2(i,j) \oplus BN_3(i,j) \oplus BN_4(i,j)) \quad (4)$$

where $Output(i,j)$ represents the output of the final merged multiscale feature fusion, $BN_2(i,j)$ represents a matrix of convolution processing with kernel size of 2, $BN_3(i,j)$ represents a matrix of convolution processing with kernel size of 3, $BN_4(i,j)$ represents a matrix of convolution processing with kernel size of 4, $\oplus$ represents the splicing operator, and $Concatenate$ represents the concatenate layer.

$$Max\_pooling(i,j) = Max_{pool\_size}(Output(i,j)) \quad (5)$$

where $Max\_pooling(i,j)$ represents the max-pooling output, and $pool\_size$ represents the size of the max-pooling kernel.

### C. Multiple Max-pooling Integration Module

The second part in stage 2 of MC-Net is a multiple max-pooling integration module. This module consists only of one convolution layer with a kernel-size of 1 × 1, but pool-sizes have different values. The main purpose of the convolution layer is to integrate the max-pooling layer output feature maps from each submodule in the encoder network, to reduce the amount of parameter computation, and to express more complex features through a small number of parameters. The max-pooling layers of different sizes retain the translation invariance of feature maps to the maximum extent, and the global context information is preserved by aggregating the feature maps from different submodules of the encoder layer network. As shown in Figure 1, number "32" in the light blue square inside this module (the "Integration Module" section) indicates that the feature maps generated by the convolution of 1 × 1 are compressed by a max-pooling layer with pool-size of 32. Similarly, the other numeric symbols "1", "2", "4", "8", and "16" are associated with the respective numbers in the encoder network. The feature maps generated by different modules in the encoder network are compressed by the max-pooling layer with different pool-size values. The Multiple Max-pooling Integration Module strategy we proposed is described in detail in Figure 2. The most important parts of all the feature maps are retained, and the feature maps are spliced by the concatenation layer. The above processing provides more prior information for the succeeding decoder network to refine the feature maps. All the equations for the integration module are as follows:

$$Con_{1\_m}(i,j) = \sum_{f=1}^{n}(X_{Max\_pooling\_m(i,j)} * W_f)(i,j) + b \quad (6)$$

where $Con_{1\_m}(i,j)$ represents a convolution with kernel size of 1, $m$ represents the output matrix of the m submodules in the encoder network, and $1\_m$ represents that the size of the convolution kernel of submodule m is 1.

$$Max\_pooling_{1\_m}(i,j) = Max_{pool\_size}(Con_{1\_m}(i,j)) \quad (7)$$

$$Integ\_Module(i,j) = Concatenate(Max\_Pooling_{1\_1}(i,j) \oplus ... \oplus BN_{1\_m}(i,j)) \quad (8)$$

where $Integ\_Module(i,j)$ represents the output of the multiple max-pooling integration module.

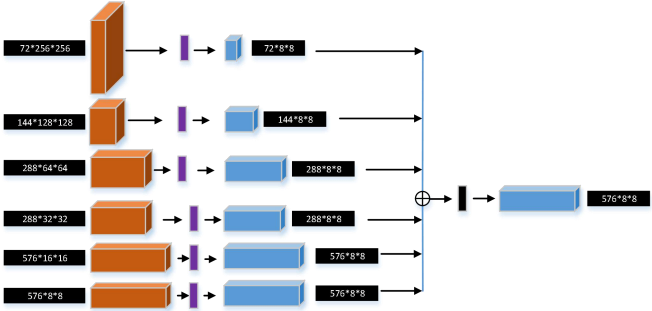

Fig. 2. Illustration of the Multiple Max-pooling Integration Module. The purple rectangle represents the max-pooling layer; The black rectangle represents the convolution layer of 1 × 1; and $\oplus$ represents fusion algorithm; The first number in the black matrix with numbers represents the number of convolution filters; the second and third numbers represent the size of the current feature maps; * represents multiplication.

*D. Decoder Network*

The main function of the decoder network is to refine the feature maps generated by the encoder network. In each submodule, we use the upsampling layer (bilinear interpolation) to directly copy the features to extend feature maps. Using the output of the upsampling layer as the input of three convolutional layers with different convolution kernel sizes, three kinds of feature maps are generated by trainable multiscale convolution. We use these three kinds of feature maps and the feature maps generated by multiscale convolution in the encoder network to perform a cross operation implemented by using the add layer (e.g., the convolution generated by the convolution kernel size of 2 in the encoder network is used as a priori information of convolution with kernel size of 3 in the decoder network).

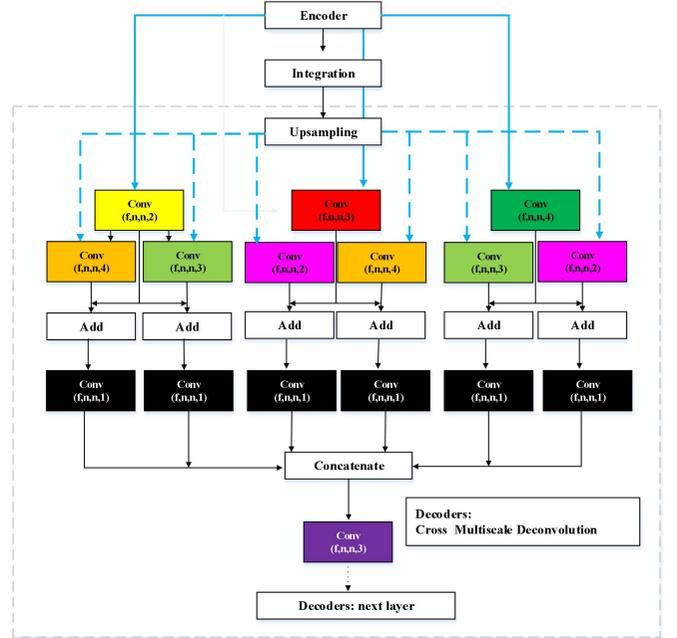

Fig. 3. Illustration of the cross multiscale deconvolution. 'Upsampling' represents the Upsampling layer; 'Conv' represents the convolution layer (f represents the number of filters, n represents the size of the feature map, and the number represents the convolution kernel size); 'Add' represents the Add layer; and ' Concatenate ' represents the Concatenate layer.

From Fig. 3, we can see that the feature maps generated by the convolution operation in the encoder is used as the prior feature information of the feature maps generated by different convolution operations in the decoder, and the generated feature maps are integrated through the convolution with a convolution kernel of 1. We integrate the information of the feature maps generated by the cross operation using the convolution kernel size of 1 and add the nonlinear activation function to the convolution, so the network can express more complex features with a small number of parameters (In the Kaggle and CHAOS datasets, the calculation went through the Relu activation function and then through the normalization calculation, but we found that it is better to calculate through normalization first and then through Relu activation function in BraTS dataset.).

And then, we splice the integrated feature maps using the concatenate layer to provide more valuable feature maps for the final classifier. Finally, we do feature integration again through convolution with convolution kernel 1, and deep feature mining through convolution with convolution kernel 3. In stage 3, we use different classification functions to classify different datasets. Equation 9 introduces another strategy proposed in this paper, namely, the cross operation on the results of different convolution kernels.

$$Decoder\_output(i,j) = Add(BN_2(i,j) \\ = \otimes BN_{Decoder\_3}(i,j)) \quad (9)$$

where $Decoder\_output(i,j)$ represents the output of the decoder network, $\otimes$ represents the addition of feature dimensions, $Add$ represents the add layer, and $BN_{Decoder\_3}(i,j)$ represents the convolution operation's result of kernel size of 3 in the submodule of the decoder network.

IV. EXPERIMENTS AND RESULTS

*A. Datasets*

We chose three datasets for the experiments; a short description of each dataset is given as follows.

**Kaggle 2018 data science bowl dataset** [38]. This dataset was provided by the 2018 Cell Segmentation Competition organized by the Booz Allen Foundation. This dataset consists of 660 cell features and provides ground truth for each single cell. First, we merged all the single-cell segmented images into one complete image. Then we expanded the size of the complete image and its corresponding ground truth to 256 × 256. (The training set to the testing set ratio is 3:2.)

**CHAOS dataset** [23]. This dataset is the Healthy Abdominal Organ Segmentation Challenge. It includes five segmentation classes with background. We focus on the segmentation of abdominal organs on MRI (T1-DUAL in phase). We selected 647 images with ground truth as the evaluation input of our model. We divided the 647 images into new training sets (three fifths of the total images) and test sets (the rest two fifths).

**BraTS 2018 dataset** [39], [17]. The training set in this dataset consists of 210 glioblastoma (HGG) and 75 lower-grade glioma (LGG) images. Since the testing set does not have ground truth data, we generate a new division of the old training set (the training set to the testing set ratio is 3:2). This dataset is multimodal and composed of four different MRI images: FLAIR, T1, T1CE and T2. The dataset is divided into four classes: edema (ground truth of 1), non-enhancing solid core (ground truth of 2), enhancing core (ground truth of 3), and the remaining images with ground truth of 0. We selected 40 of 155 slice images from each case as input for model training and testing. Because the size of the original slice image is 240 × 240, we padded the original image with zeros to make the image size 256x256.

*B. Experimental evaluation criteria*

Because Kaggle dataset is mainly used to solve the problem of binary classification, we use six evaluation indicators, namely, Accuracy (*Accuracy, Acc*), Precision (*P*), F-Measure (*F*), Sensitivity (*Sen*), Specificity (*Spec*) and the Dice score (*Dice*) to evaluate the performance of a model. They are all expressed as percentage in the experimental result tables. Using $TN$ to represent the number of correct negative sample classifications, $FN$ to represent the number incorrect negative sample classifications, $FP$ to represent the number of incorrect positive sample classifications, and $TP$ to represent the number of correct positive sample classifications, the indicators are defined as follows:

$$P = TP/(TP+FP) \quad (10)$$

$$F = (2P*R)/(P+R) \quad (11)$$

$$Accuracy = (TP+TN)/(TP+FN+FP+TN) \quad (12)$$

$$Sen = TP/(TP+FN) \quad (13)$$

$$Spec = TN/(TN+FP) \quad (14)$$

$$Dice = (2*TN)/(2*TN+FN+FP) \quad (15)$$

The CHAOS dataset consists of four classes including Liver, Kidney R, Kidney L and Spleen. We will evaluate the accuracy of each of its classes separately, as well as their sensitivity (Sen), specificity (Spec) and dice score (Dice).

The BraTS dataset has its own evaluation criteria [8]. This dataset evaluates WT (including all four tumor structures), ET (including all tumor structures except "edema"), and TC (only containing the "enhancing core" structures that are unique to high-grade cases) using Dice*, Sens*, and Spec* evaluation criteria. The metrics are calculated using the following equations:

$$Dice(M,N)^* = |M_1 \wedge N_1|/((|M_1|+|N_1|)/2) \quad (16)$$

$$Sens(M,N)^* = |M_1 \wedge N_1|/|N_1| \quad (17)$$

$$Spec(M,N)^* = |M_0 \wedge N_0|/|N_0| \quad (18)$$

where $M_0$ and $N_0$ are pixels where $M=0$ and $N=0$, respectively. Operator $\wedge$ is the logical AND operator, $||$ is the size of the set (i.e., the number of pixels belonging to it), and $M_1$ and $N_1$ are the sets of pixels where $M=1$ and $N=1$. The value of 1 represents all the current evaluation objects, and 0 represents the remaining objects.

*C. Experimental parameter setting*

The server graphics card we use is NVIDIA Tesla V100 16 GB. We used the Adam optimizer because it is not only simple and efficient, but also is not affected by gradient expansion during parameter updating. The learning setting is 0.00001. For the binary classification problem, we use the sigmoid activation function as the classification result and binary cross-entropy as the loss function to calculate. For the multi-classification problem, we use the softmax activation function as the classification result and categorical cross-entropy as the loss function to calculate. The number of fitters in the MC-Net model is [72, 144, 288, 288, 576] in the Encoder Module. In Integration Module, they are [72, 144, 288, 288, 576, 576]; In the Decoder Module are [576, 288, 288, 144, 72]. In order to make each model finally fit on each dataset, we iterated 400 times for each model on Kaggle dataset and CHAOS dataset, and 200 times on BraTS dataset.

We conducted a comprehensive evaluation of the model using the three datasets. All training and testing data have not been preprocessed. All the models compared include some classical and the state-of-the-art end-to-end models which have been extraordinarily influential. The experimental results are shown in tables I, II and III, and the visualization results after segmentation are shown in Figures 4, 5 and 6.

### D. Model performance

TABLE I SEGMENTATION RESULTS FOR THE KAGGLE 2018 DATA SCIENCE BOWL DATASET OBTAINED WITH EACH MODEL

| Dataset | Model | Acc | P | F | Sen | Spec | Dice |
|---|---|---|---|---|---|---|---|
| Kaggle | FCNs[31] | - | - | - | - | - | - |
| | DecovNet[33] | 89.72 | 64.54 | 53.26 | 46.62 | 95.91 | 93.61 |
| | PSPNet[20] | 93.59 | 74.94 | 69.17 | 65.72 | 97.21 | 95.98 |
| | SegNet[34] | 93.67 | 83.71 | 61.77 | 70.25 | 98.28 | 96.07 |
| | U-Net[35] | 97.21 | 87.25 | 87.18 | 88.18 | 98.31 | 98.2 |
| | HyperDenseNet[40] | 93.85 | 82.18 | 76.32 | 77.13 | 97.7 | 96.1 |
| | Espnetv2[43] | 96.8 | 86.59 | 86.3 | 87.14 | 98.06 | 97.97 |
| | Denseaspp[42] | 93.95 | 76.77 | 69.91 | 65.51 | 97.22 | 96.23 |
| | MS-Dual[41] | 98.08 | 91.44 | 91.45 | 92.92 | 98.66 | 98.66 |
| | MC-Net | **98.16** | **91.49** | **91.71** | **93.21** | **98.80** | **98.81** |

TABLE II SEGMENTATION RESULTS FOR THE CHAOS DATASET OBTAINED WITH EACH MODEL

| Dataset | Model | Liver | Kidney L | Kidney R | Spleen | Sen* | Spec* | Dice* |
|---|---|---|---|---|---|---|---|---|
| CHAOS | FCNs[31] | - | - | - | - | - | - | - |
| | DecovNet[33] | 68.07 | 39.95 | 44.1 | 37.32 | 67.56 | 60.28 | 97.53 |
| | PSPNet[20] | 62.44 | 40.06 | 30.78 | 45.29 | 68.32 | 56.04 | 97.72 |
| | SegNet[34] | 58.27 | 24.88 | 22.97 | 24.33 | 62.33 | 48.75 | 97.43 |
| | U-Net[35] | 61.51 | - | - | - | 50.45 | 44.38 | 97.3 |
| | HyperDenseNet[40] | - | - | - | - | - | - | - |
| | Espnetv2[43] | 66.81 | 55.72 | 57.98 | 59.47 | 75.66 | 64.17 | 98.2 |
| | Denseaspp[42] | 73.55 | 60.53 | 62.77 | 62.5 | 77.47 | 70.33 | 98.22 |
| | MS-Dual[41] | 92.46 | 87.96 | **88.01** | 78.61 | 88.34 | 89.95 | 99.31 |
| | MC-Net | **93.74** | **89.03** | 87.58 | **92.05** | **93.45** | **92.65** | **99.48** |

TABLE III SEGMENTATION RESULTS FOR THE BRATS 2018 DATASET OBTAINED WITH EACH MODEL

| Dataset | Model | Dice* | | | Sens* | | | Spec* | | |
|---|---|---|---|---|---|---|---|---|---|---|
| | | ET | WT | TC | ET | WT | TC | ET | WT | TC |
| BraTS | FCNs[31] | - | - | - | - | - | - | - | - | - |
| | DecovNet[33] | 14.51 | 13.63 | 2.06 | 36.19 | 34.02 | 12.57 | 86.7 | 87.25 | 86.03 |
| | PSPNet[20] | 48.83 | 51.73 | 50.55 | 52.03 | 52.9 | 52.59 | 96.74 | 97.13 | 95.97 |
| | SegNet[34] | 13.93 | 17.24 | 17.11 | 40.99 | 41.74 | 31.04 | 86.93 | 87.33 | 86.22 |
| | U-Net[35] | 63.02 | 61.32 | 57.41 | 60.84 | 58.78 | 59.37 | 99.3 | 99.53 | 98.73 |
| | HyperDenseNet[40] | - | - | - | - | - | - | - | - | - |
| | Espnetv2[43] | 80.75 | 80.61 | 79.95 | 81.14 | 79.43 | 80.26 | 99.54 | **99.78** | 99.17 |
| | Denseaspp[42] | 76.95 | 77.57 | 75.63 | 77.03 | 76.38 | 79.18 | 99.43 | 99.71 | 98.98 |
| | MS-Dual[41] | 72.83 | 70.89 | 74.3 | 73.55 | 70.51 | 73.17 | 99.29 | 99.70 | 98.65 |
| | MC-Net | **83.02** | **82.69** | **84.05** | **83.38** | **82.12** | **84.87** | **99.55** | 99.77 | **99.22** |

Table I shows in detail the experimental results of binary classification by each model on the Kaggle dataset. The values of *P* and *Sen* in the experimental results clearly indicate that the MC-Net model is more effective in predicting positive samples. The Kaggle dataset is characterized by an uneven distribution, and a large difference in shape and cell size between positive and negative samples. Therefore, it is necessary to predict the feature maps, which are more representative if they are generated by a deeper network. An MC-Net with 54 layers achieves the best experimental results. The decoder network structure of the FCNs model is relatively simple, and a substantial amount of useful information is lost in the process of decoding. As a result, the applicability of feature maps provided for prediction in the Kaggle dataset is poor. The DecovNet model obtains a more refined feature map by improving the decoder network. The PSPNet model retains more feature maps and provides more prior knowledge for the subsequent decoder network; additionally, a variety of pooling layers retain different contextual semantic information, which leads to the improvement of experimental results. Because the decoder network of the model only uses the upsampling layer, the generated feature maps perform poorly; in its pyramid pooling strategy, except for the last submodule in the encoder network, the a priori information of other submodules and the original image is missing. The feature mining ability of the SegNet model is weak, which leads to the poor experimental result of this dataset. The advantage of the U-Net model is that the feature graph in the encoder network is spliced with the refined feature graph of the decoder network, which provides more original feature maps for the classifier and improves the classification results. In some of the state-of-the-art models, because the total amount of operational parameters of HyperDenseNet is large, compared with the previous nine-layer iteration of its model, we only use five layers, and it may have an impact on the final result. Denseaspp does not perform very well on Kaggle dataset because it is specially designed for multi-scale samples in large targets. Both Espnetv2 and MS-Dual use multiple inputs to retain more deep features, so they get better results. In the MC-Net model, the encoder network retains a large number of prior feature maps. The multiple max-pooling integration module retains contextual semantic information for each submodule and integrates information by introducing convolution with a kernel size of 1. The decoder network refines the feature map through cross multiscale deconvolution and fuses the prior feature map generated by each submodule in the encoder network. The decoder network provides a more representative feature map for the final classification. From the experimental results in Table I, it can be seen that the MC-Net model achieves a better segmentation effect on the Kaggle 2018 data science bowl dataset.

Table II shows the experimental results of each model on the CHAOS dataset. It shows that the learning ability and fitting performance of FCNs and HyperDenseNet models are weak and these two models do not learn useful features. U-Net model only has a certain learning ability for Liver class, but poor learning ability for other classes. We can clearly see that the model experimental results of some state-of-the-art are better than the classical model experimental results, and we can see that the experimental results of DecovNet are better than other classical models, which proves the excellent performance of deconvolution. From Table II, we can clearly see that the classification results of our proposed MC-Net model about the four classes are better than other models. In particular, the segmentation effects of Liver, Kidney L and Spleen are 93.74%, 89.09% and 92.05%, respectively, which are the best among all the models. At the same time, the other three evaluation indicators of our proposed model MC-Net, namely Sen, Spec and Dice, are obviously better than those of other models, which shows the feasibility and robustness of our proposed model.

Table III shows the test results for the BraTS dataset, which presents a multiclass segmentation task. The structure and content of this dataset are more complex than those of the other two datasets and necessitate using four different types of MRI medical images as input. The experimental results show that the descriptive ability of feature maps generated by the FCNs model is very poor. DecovNet and SegNet have a certain descriptive ability to generate more detailed feature maps through the decoder network. PSPNet retains a substantial amount of prior knowledge to provide a more descriptive feature map for the final classification. In the decoder network, each submodule of the U-Net combines the prior characteristic diagram of each submodule of the corresponding encoder network to generate a feature map with a strong descriptive ability. It is clear from Table III that more prior feature maps are needed for complex data in order to achieve better experimental results. The operation parameters of layer 9 of HyperDenseNet model are too large, and the number of layers is too small, so the learning ability is weak and the experimental results are poor. On the whole, the learning ability of these state-of-the-art models is better than that of traditional models. Among them, the prediction ability of Espnetv2 model for enhancing core is slightly higher than that of our proposed MC-Net model. In other experimental results, the results of our proposed MC-Net model are much better than those of other models. The experimental results show that our model can better solve the task of multi-modal image segmentation. Specifically, the MC-Net model surpasses all other models according to all evaluation indexes.

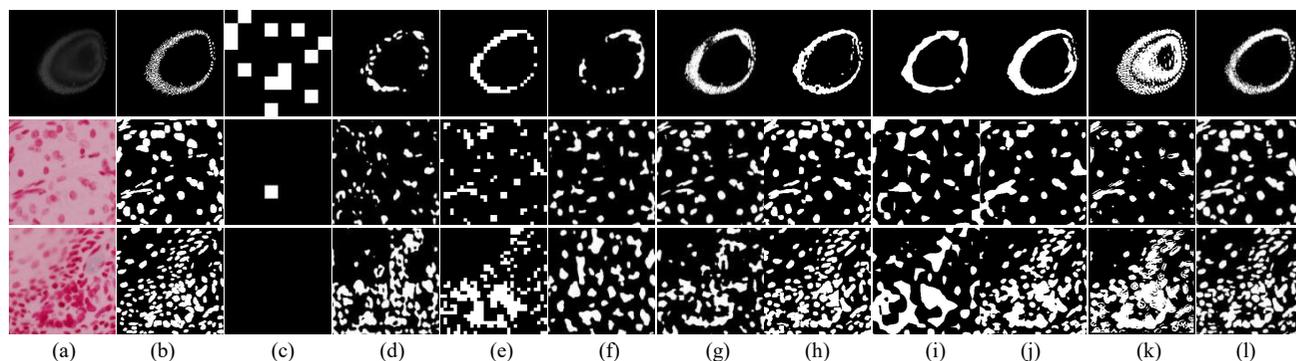

(a) (b) (c) (d) (e) (f) (g) (h) (i) (j) (k) (l)
Fig. 4. Segmentation results of each model for the Kaggle 2018 data science bowl dataset. (a) Original image. (b) Ground truth. (c) FCNs. (d) DecovNet. (e) PSPNet. (f) SegNet. (g) U-Net. (h) MS-Dual.(i) Denseaspp. (j) Espnetv2. (k) HyperDenseNet. (l) MC-Net.

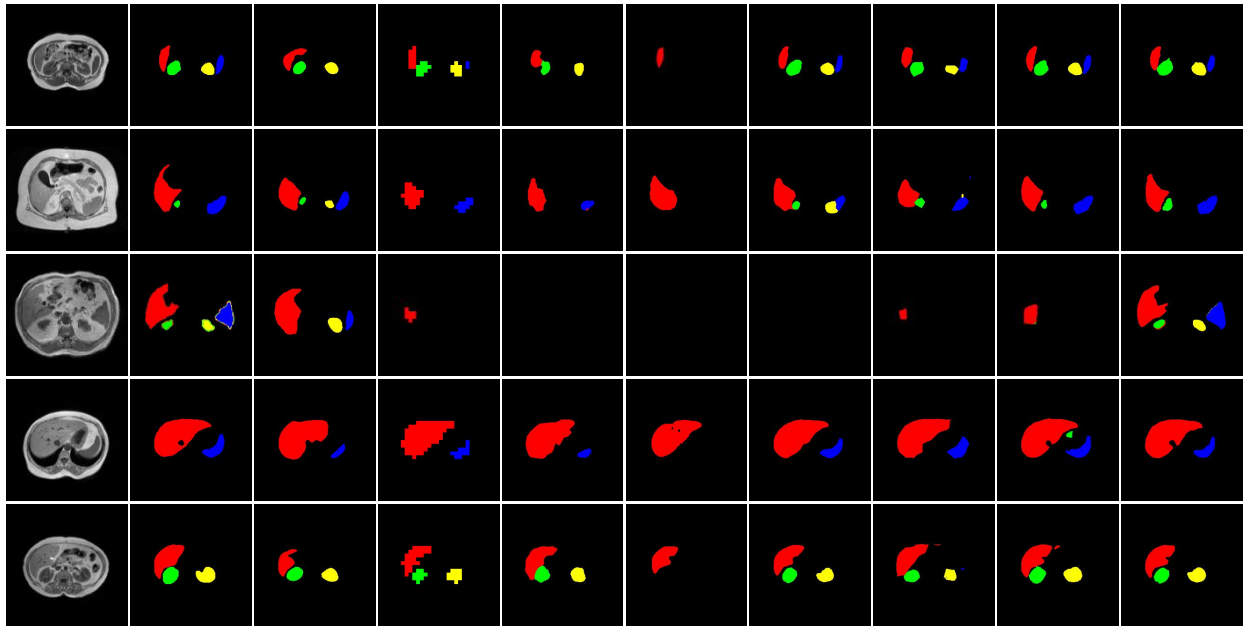

Fig. 5. Segmentation results of each model for the CHAOS dataset. (a) Original image. (b) Ground truth. (c) DecovNet. (d) PSPNet. (e) SegNet. (f) U-Net. (g) Espnetv2. (h) Denseaspp. (i)MS-Dual. (j) MC-Net. Colors indicate liver(red), Kidney L (green), Kidney R(yellow), and Spleen (blue).

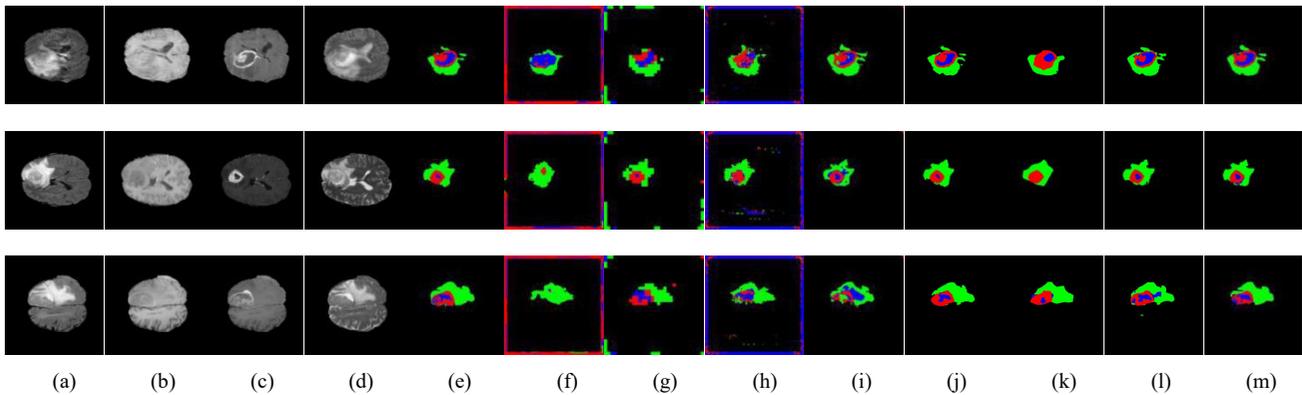

(a)    (b)    (c)    (d)    (e)    (f)    (g)    (h)    (i)    (j)    (k)    (l)    (m)

Fig. 6. Segmentation results of each model for the BraTS 2018 dataset. (a) Flair. (b) T1. (c) T1ce. (d) T2.(e) Ground truth. (f) DecovNet. (g) PSPNet. (h) SegNet. (i) U-Net. (j) Espnetv2. (k) Denseaspp. (l)MS-Dual.(m) MC-Net. Colors indicate edema (green), non-enhancing solid core (red), and enhancing core (blue).

According to Figure 4, the segmentation targets in the three test samples are irregular and small. From the visualization results, we can clearly see that the FCNs algorithm has poor fitting ability and does not use deconvolution, resulting in poor learning results and square segmentation results. The feature learning ability of the three algorithms of DecovNet, PSPNet, SegNet is limited. From the experimental results, it is noted that some of the segmentation targets are lost, for instance, PSPNet produces some square segmentation results which are not smooth, because it does not use deconvolution. Among the four state-of-the-art models, the Denseaspp model is not effective in dealing with the data of small targets, which leads to poor boundary separation of small targets. The other three models all have the ability to deal with small targets, and the Espnetv2 model has the best effect. Because U-Net integrates the feature maps from the encoder with the decoder network to obtain more prior information in the classification, it achieves higher prediction accuracy than other traditional models. Our proposed model not only obtains a large number of prior feature graphs in the encoder network and multiple max-pooling integration module, but also uses the cross multiscale deconvolution operation to generate a large number of feature maps in the decoder network. It produces better prediction results for small targets and irregular targets.

Figure 5 shows the results of CHAOS dataset segmentation. From the first group of experimental results, it is clear that the results from some state-of-the-art models are better than those from the classical models. The second group of experimental results show that the effect of MS-Dual and MC-Net model in segmenting Kidney R category are the best. In other models, either the feature maps mining are not thorough enough or the feature maps are lost, which leads to the omission of the segmentation target. The third group of experimental results demonstrated that our proposed MC-Net model can better learn the deep characteristics of the data, and the effect is much better than the learning results of other models. The fourth group of experimental results show that the edge processing ability of our proposed algorithm is better than that of other models. The fifth group of experimental results show that the MC-Net model has a better denoising effect than some of the latest models, indicating that our proposed model plays a better role in information integration.

Figure 6 shows the results of segmentation of a multiclass dataset. The figures clearly show that the MC-Net model is superior to other models in classification, image denoising, edge processing, etc. The FCNs models do not learn any useful features in this dataset, and the output segmentation results are all black; hence, these models are not included in Figure 6. The prediction results of DecovNet, PSPNet and SegNet are poor. In addition, there is a substantial amount of noise in the resulting segmentation figure produced by the SegNet model; the predictive ability of the DecovNet model is weaker than those of the other two models, and the overall prediction result of the PSPNet model is better, but the edge processing ability of the model is weak, which results in blurry edges.

### E. Evaluation of the number of model parameters and layers

The number of layers and parameters of a model play a key role in feature learning. To further analyze the performance of the MC-Net model, we study the number of layers and parameters of the model and evaluate the model by the error rate (1 - Accuracy rate). Table IV shows the experimental results for various numbers of layers in the model, and the number of sub-modules (including encoder network, decoder network and multiple max-pooling integration module). Table V shows the experimental results for various numbers of parameters.

TABLE IV Experimental results for various numbers of layers in the MC-Net Model

| Model | Kaggle | CHAOs | BraTS |
|---|---|---|---|
| MC-Net(2) | 2.56 | 1.69 | 1.73 |
| MC-Net(3) | 2.11 | 0.68 | 1.24 |
| MC-Net(4) | 2.02 | 0.55 | 0.97 |
| MC-Net(5) | 1.89 | **0.52** | **0.78** |

It can be seen from Table IV that with the increase of the number of model layers, the error rate of our proposed model MC-Net in learning each dataset has been decreased. From the results of Kaggle data sets, it is obvious that when the model has five sub-modules, the error rate is the lowest. With the decrease of the number of model sub-modules, the error rate increases gradually, but the gap is small, which shows that our proposed model can learn useful information in the image even if the number of modules is reduced. In the CHAOs dataset, the error rate of the model composed of five sub-modules is close to that of the model composed of four sub-modules, but the error rate of the model composed of two sub-modules is higher. It is because the CHAOs data set is a multi-classification task, when the model is composed of only two sub-modules, it can not well learn the deep features of the CHAOs dataset, resulting in a significant degradation in the results. The error rate of the experimental results of MC-Net for CHAOs dataset is only 0.52%, which further proves that our proposed model has better learning features. The BraTS dataset is composed of four different modes, and the background pixels account for 97.3% of the total pixels. When the number of sub-modules of the model get smaller and smaller, the learning effect will become worse and worse. The complexity of the dataset is high, but the error rate of MC-Net learning is only 0.78%. The results using the above three datasets demonstrated the feasibility of our proposed model.

TABLE V Comparison of experimental results of various models with the same parameters

| Model | Kaggle | CHAOs | BraTS | Number of parameters |
|---|---|---|---|---|
| DecovNet[33] | 13.43 | 3.73 | 15.33 | 10 M |
| PSPNet[20] | 6.82 | 2.63 | 7.8 | 10 M |
| SegNet[34] | 6.66 | 4.74 | 15.02 | 10 M |
| U-Net[35] | 3.23 | - | 2.96 | 7.7 M |
| Espnetv2[43] | 3.58 | 1.71 | 0.91 | 4 M/10M |
| Denseaspp[42] | 6.38 | 1.68 | 1.05 | 10 M |
| MS-Dual[41] | 2.02 | 0.72 | 1.32 | 6.3 M |
| MC-Net | **1.89** | **0.54** | **0.81** | 6.8 M |

In theory, as the number of parameters increases, we can retain more features and eventually obtain better results. Hence, as shown in Table V, we use the parameters of the SegNet model as the benchmark. If the scale of the image is 256 × 256, the total number of parameters of the SegNet model is approximately 10 M. To avoid changing the structure of the model, we only modify the number of fitter in each layer of the model. When the size of the input image is 256 * 256 * 1, the total number of parameters of Espnetv2 is 4,777,180, but when the size of the input image is 256 * 256 * 4, the total number of parameters of Espnetv2 is 74,197,943, so for the Espnetv2 model, we use 4M parameters on Kaggle and CHAOs datasets, and 10M parameters on BraTS dataset. The experimental results of these three datasets show that our proposed MC-Net model still achieves the lowest error rate when the parameters are reduced.

*F. Strategy evaluation*

To further assess the effectiveness of strategies in the MC-Net model, we use three datasets to evaluate different strategies: not using the cross multiscale deconvolution and not using the integration module which is therefore called "No-strategy", the strategy 1 which is referred to as not using the cross multiscale deconvolution but using the integration module, and the strategy 2 which is referred to as not using the integration module but using the cross multiscale deconvolution. A line chart is used to further evaluate the feasibility of our proposed strategies. Tables VI, VII and VIII show the specific results of each strategy on the three datasets.

TABLE VI Segmentation results for the Kaggle 2018 data science bowl dataset obtained with various strategies

| Dataset | Model | Acc | P | F | Sen | Spec | Dice |
|---|---|---|---|---|---|---|---|
| Kaggle | No-strategy | 97.71 | 88.32 | 89.17 | 91.48 | 98.04 | 98.11 |
|  | MC-Net-strategy1 | 97.79 | 89.98 | 90.25 | 91.53 | 98.35 | 98.37 |
|  | MC-Net-strategy2 | 97.87 | 90.31 | 90.56 | 91.88 | 98.46 | 98.44 |
|  | MC-Net | **98.16** | **91.49** | **91.71** | **93.21** | **98.80** | **98.81** |

TABLE VII Segmentation results for the CHAOS dataset obtained with various strategies

| Dataset | Model | Liver | Kidney L | Kidney R | Spleen | Sen* | Spec* | Dice* |
|---|---|---|---|---|---|---|---|---|
| CHAOs | No-strategy | 90.3 | 76.05 | 75.56 | 85.28 | 88.31 | 88.2 | 99.03 |
|  | MC-Net-strategy1 | 90.73 | 76.52 | 75.8 | 85.73 | 88.91 | 88.68 | 99.05 |
|  | MC-Net-strategy2 | 90.78 | 77.27 | 77.54 | 91.02 | 90.29 | 88.93 | 99.16 |
|  | MC-Net | **93.74** | **89.03** | **87.58** | **92.05** | **93.45** | **92.65** | **99.48** |

TABLE VIII Segmentation results for the BraTS 2018 dataset obtained with various strategies

| | Model | Dice* | | | Sens* | | | Spec* | | |
|---|---|---|---|---|---|---|---|---|---|---|
| | | ET | WT | TC | ET | WT | TC | ET | WT | TC |
| BraTS | No-strategy | 74.83 | 72.9 | 76.3 | 75.55 | 72.51 | 75.17 | 99.29 | 99..69 | 98.8 |
| | MC-Net-strategy1 | 79.79 | 78.94 | 80.46 | 80.38 | 78.31 | 79.91 | 99.43 | 99.75 | 99.02 |
| | MC-Net-strategy2 | 76.92 | 75.5 | 76.83 | 77.94 | 74.7 | 77.68 | 99.37 | 99.74 | 98.91 |
| | MC-Net | **83.02** | **82.69** | **84.05** | **83.38** | **82.12** | **84.87** | **99.55** | **99.77** | **99.22** |

From the three groups of experiments in Tables VI, VII and VIII, we can see that better experimental results can be obtained by using the strategies. The results of each strategy are effective, which proves that the strategy proposed by us is feasible. When the MC-Net model uses one or more strategies, the experimental results are significantly higher than those without strategies. The experimental results of Kaggle dataset and CHAOs dataset show that the cross multiscale deconvolution strategy is better than the integration module strategy, but the integration module strategy is better than the cross multiscale deconvolution strategy in BraST data. We speculate that the integration module strategy is more suitable for multimodal fusion.

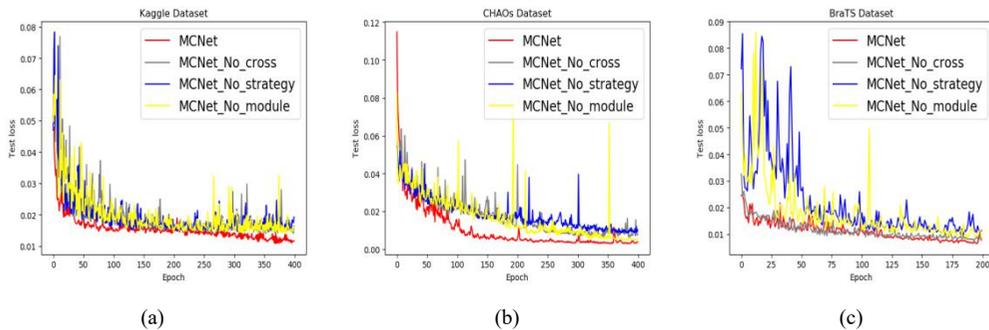

Fig. 7. (a) Resulting loss value of the testing and training sets for the Kaggle 2018 data science bowl dataset. (b) Resulting loss value of the testing and training sets for the CHAOs dataset. (c) Resulting loss value of the testing and training sets for the BraTS 2018 dataset.

The three line charts in Fig. 7 represent the loss results of MC-Net's testing sets in three datasets.

Fig. 7(a) shows the experimental results for the Kaggle 2018 data science bowl dataset. It is obvious that using two strategies at the same time makes the learning ability of the model more stable. Considering the experimental results for the Kaggle dataset, Fig. 7(b) shows that MC-Net is also suitable for multi-classified data sets and has stronger fitting ability. Fig. 7(c) shows the results of the experiment on the BraTS dataset. The fitting performance of cross multiscale deconvolution strategy on multimodal data sets is much higher than that of integration module strategy. The results for the above three datasets clearly show that the experimental results are the best if the MC-Net model uses two strategies simultaneously.

V. CONCLUSIONS

The experimental results on the aforementioned three datasets show that the strategy of using the multiple max-pooling integration module can retain more contextual semantic information in the feature maps of each submodule and in the original image. The strategy of using cross multiscale deconvolution can provide different prior information for the feature maps generated by different convolutions and therefore improve the refining ability of feature maps. Using the two strategies in developing the MC-Net model, we obtain better medical image segmentation results using fewer parameters. We hope that the detailed analysis of the MC-Net model can help medical image segmentation and provide the reference for the development of related technologies.